# Smart Content Recognition from Images Using a Mixture of Convolutional Neural Networks[*]

Tee Connie[*], Mundher Al-Shabi[*], and Michael Goh

Faculty of Information Science and Technology, Multimedia University, Melaka, Malaysia

**Abstract.** With rapid development of the Internet, web contents become huge. Most of the websites are publicly available, and anyone can access the contents from anywhere such as workplace, home and even schools. Nevertheless, not all the web contents are appropriate for all users, especially children. An example of these contents is pornography images which should be restricted to certain age group. Besides, these images are not safe for work (NSFW) in which employees should not be seen accessing such contents during work. Recently, convolutional neural networks have been successfully applied to many computer vision problems. Inspired by these successes, we propose a mixture of convolutional neural networks for adult content recognition. Unlike other works, our method is formulated on a weighted sum of multiple deep neural network models. The weights of each CNN models are expressed as a linear regression problem learned using Ordinary Least Squares (OLS). Experimental results demonstrate that the proposed model outperforms both single CNN model and the average sum of CNN models in adult content recognition.

## 1    Introduction

The number of Internet users increases rapidly since the introduction of World Wide Web (WWW) in 1991. With the growth of Internet users, the content of the Internet becomes huge. However, some contents such as adult content are not appropriate for all users. Filtering websites and restricting access to adult images are significant problems which researchers have been trying to solve for decades. Different methods have been introduced to block or restrict access to adult websites such as IP address blocking, text filtering, and image filtering. The Internet Protocol (IP) address blocking bans the adult content from being accessed by certain users. This technique works by maintaining a list of IPs or Domain Name Servers (DNS) addresses of such non-appropriate websites. For each request, an application agent compares the requested website IP address or DNS with the restricted list. The request is denied if the two addresses match, and approved otherwise. This method requires manual keeping and maintenance of the restricted list IPs, which is difficult as the number of the adult content websites grows or some websites change their addresses regularly.

---

[*]   These authors contributed equally to this work

Filtering by text is the most popular method to block access to adult content websites. The text filtering method blocks the access to a website if it contains at least one of the restricted words. Another approach is to use a machine learning algorithm to find the restricted words. Sometimes, instead of using the machine learning technique to extract keywords, a classification model is used directly to decide whether the requested webpage is safe [7]. Nonetheless, the text blocking method only understands texts, and it cannot work with images. This problem arises when the webpage does not contain the restricted keywords or does not contain text at all. Worse still, it may block safe webpages such as a medical webpage as it contains some restricted keywords.

Another blocking method uses image filtering [1, 9, 11]. This method works directly on the images, trying to detect if the image contains adult content. Detection directly from images is favorable as it does not require a list of IPs and is scalable to new websites, and is not sensitive to certain keywords. However, detecting adult content from images requires a complex model as the images have different illuminations, positions, backgrounds, resolutions or poses. In addition, the image may contain part of the human body, or the person in the image may be partially dressed.

In this paper, we seek to automatically recognize adult content from images using a mixture of convolutional neural networks (CNNs). Fig. 1 shows the architecture of the proposed model in which eight CNNs models, followed by Fully Connected (FC) layers, are used to vote for the possible class of the image. Each model conforms to the same architecture with different weights computed using Ordinary Least Square (OLS). Usually, the training time of the deep CNN is very long. We present a solution to create eight models from a single architecture during training. A checkpoint is set to identify and pick the eight most-performing models during the training session. The solution selects the most optimal model to improve accuracy and helps reduce the training time drastically.

The contributions of this paper are as follows: 1) constructing a mixture of multiple deep CNNs at no extra cost; 2) assigning different weights to every model by applying OLS on all the model's output predictions

## 2    Related works

The methods of recognizing adult content images can be divided into four categories: color-based, shape information-based, local features-based, and deep-learning-based.

The first approach analyzes the images based on skin color. This method classifies a region of pixels as either skin or non-skin. The skin color can be detected manually using a color range [1], computed color histograms [4], or parametric color distribution functions [3]. Once a skin color model of the image has been defined, the adult image can be detected by a simple skin color histogram threshold, or by passing the statistics of the skin information to a classifier [11].

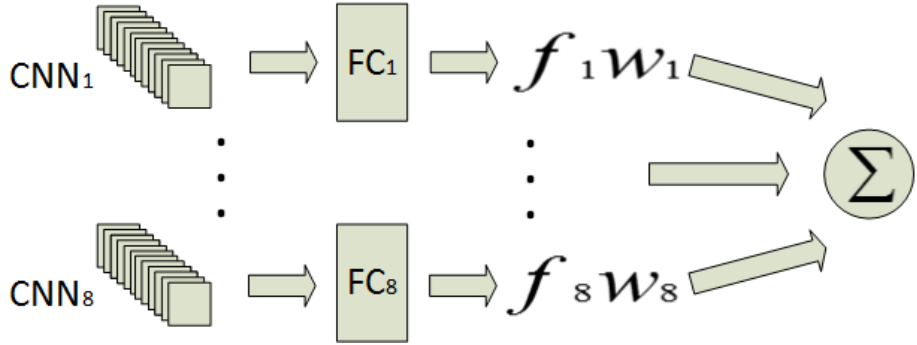

**Fig. 1.** The mixture of CNNs in which the eight CNNs models are combined linearly.

Often, the skin areas contain some shape information such as ellipses or color compactness in some parts of the human body. The structure of a group of skin color regions is analyzed to see how they are connected. Several methods have been proposed to detect the shape features such as contour-based features [1] where the outlines of the skin region are extracted and used as a feature, Hu and Zernike moments of the skin distribution [13], and Geometric constraints which model the human body geometry [2].

The third approach based on local features is inspired by the success of local features in other image recognition problems. Scale Invariant Feature Transform (SIFT) was used in conjunction with the bag of words to recognize adult images in [9]. The resulting features were trained using linear Support Vector Machine (SVM). Another local features called Probabilistic Latent Semantic Analysis was proposed in [8] to convert the image into a certain number of topics for adult content recognition.

The fourth and the most recent type of image content recognition technique is the use of deep learning approach. Moustafa [10] adopted AlexNet-based classifier [6] and the GoogLeNet [12] model architecture. Both models were treated as consultants in an ensemble classifier. Simple weighted average with equal weights was used to combine the predictions from the two models. Zhou, Kailong, et al. [14] proposed Another model based on deep learning. A pre-trained caffenet model was developed and the last two layers were fine-tuned with adult images dataset.

## 3 The Proposed Mixture of CNN Model

The proposed network contains six convolutional layers followed by two fully-connected layers as shown in Fig. 2. The number of filters in each convolutional layer is monotonically increasing from 16 to 128. A 2x2 Max-Pooling is inserted after each of the first two layers and after the fourth and the sixth convolutional layer. The size of each filter is 3x3 with two-pixel stride. To prevent the network from shrinking after each convolution, one pixel is added to each row and column before passing the image or the feature to the next convolution. After the six convolutional layers, the features bank is flattened and is passed to a fully-connected layer with 128 neurons. The output layer which only contains one neuron is placed after the first fully-connected layer, and before the sigmoid activation function. Except for the last layer, a rectifier linear unit (Relu) is used as the activation function which is less prone to vanishing gradient as the network grows. Another reason to adapt Relu is that it operates very fast as only a simple $max$ function is used.

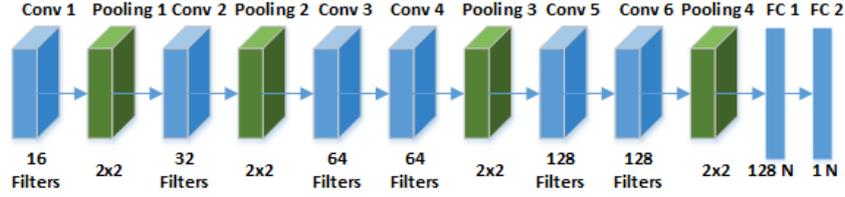

**Fig. 2.** The architecture of deep convolutional neural network model

To prevent the network from over-fitting the data, two regularization techniques have been applied. The first technique applies $L2$ weight decay with 0.01 on the first fully-connected layer. Dropout is also used to prevent over-fitting. Dropout works by randomly zeroing some of the neurons output at training to make the network more robust to small changes. Dropout is placed directly after each Max-Pooling, and also after the first fully-connected layer. The probabilities of these four dropouts are set to 0.1, 0.2, 0.3, 0.4, and 0.4, respectively.

The network is trained for 300 epochs, with each epoch consisting of multiple batches optimized with Adam [5]. The batch size is 128 and is trained with the cross-entropy loss function. As adult image recognition is a binary problem in which the output can be either positive or negative, the binary cross-entropy is used

$$L(f, y) = -\sum f \log y + (1 - y) \log(1 - f) \qquad (1)$$

where $f$ is the predicted value and $y$ is the true value.

Generally, the training time of deep CNN is very long. To alleviate this problem, we introduce a way to extract eight sub-models with different weights in a single training session.

---

**Algorithm 1** Generating Best Eight Performing Models

---

**Input:** Training data, $X$, validation data, $V$,
set of epochs $\{1,2,…,300\}$, $Q$
**Output:** The top 8 models, $top8$
**Procedure:**
   $checkpoints \leftarrow$ empty list
   $model \leftarrow$ *Deep CNN model with random weights*
   $a \leftarrow -\infty$
   **for each** epoch $\in Q$
     $model \leftarrow$ train $model$ *on X* using Adam

---

```
    accuracy ← validate model on V
    if accuracy > a
        a ← accuracy
        add model to the checkpoints list
    end
    top8 ← top 8 models in checkpoints
    return top8
```

All the eight models are validated on the validation set, and a $N \times 8$ matrix is constructed from the outputs

$$\mathbf{F} = \begin{bmatrix} f_{11} & \cdots & f_{81} \\ \vdots & \ddots & \vdots \\ f_{1N} & \cdots & f_{8N} \end{bmatrix} \quad (2)$$

where $N$ is the number samples in the validation set, and $f_{in}$ is the predicted output of the $i$-th model of the $N$ samples. The eight models combined linearly as,

$$z(w) = \begin{cases} 1, & f_1 w_1 + f_2 w_2 + \cdots + f_8 w_8 > 0 \\ 0, & Otherwise \end{cases} \quad (3)$$

Finding the unknown weights vector $\mathbf{W} = (w_1, \dots, w_8)$ that minimizes $min_{\mathbf{W}}(\mathbf{Y} - \mathbf{Z}(\mathbf{W}))^2$ is possible as long as $N >> 8$. This problem is solved using Ordinary Least Squares (OLS) as,

$$\mathbf{Z} = \begin{bmatrix} f_{11} & \cdots & f_{81} \\ \vdots & \ddots & \vdots \\ f_{1N} & \cdots & f_{8N} \end{bmatrix} \begin{bmatrix} w_1 \\ \vdots \\ w_8 \end{bmatrix} \quad (4)$$

By taking the derivative of $(\mathbf{Y} - \mathbf{Z}(\mathbf{W}))^2$ setting it equal to zero with respect to $\mathbf{W}$

$$\frac{d}{dW}(\mathbf{Y} - \mathbf{Z}(\mathbf{W}))^2 = 0 \quad (5)$$

Finally, we rearrange the equation and solve it for $\mathbf{W}$

$$\mathbf{W} = (\mathbf{F}^\mathrm{T}\mathbf{F})^{-1}\mathbf{F}^\mathrm{T}\mathbf{Y} \quad (6)$$

Usually, a model with a higher accuracy will get a higher weight than a model with lower accuracy.

## 4    Dataset

Due to the nature of the problem, there is no recognizable public dataset on adult content recognition. In this research, we collected the data manually from the Internet yielding 41,154 adult images. For negative images, images from the ILSVRC-2013 [15] test set were used. These data are separated into training, validation, and testing sets as shown in Table 1.

Each image is resized, centered, and cropped from the middle region to $128 \times 128$ pixels in RGB format. After that, all the images are normalized and mean subtracted. RGB images are fed as input to the network as they allow the first convolutional layer to extract the skin color information while the rest of the networks extract high-level features based on body shape or textures.

## 5 Experimental results

The CNN model is trained on 56,914 images and the validation set is used to pick the best eight models. In order to increase the generality of the model, the training data is augmented by flipping each image horizontally. This increases the total training images to 113,828.

**Table 1.** Dataset distribution

|            | Positive(Porn) | Negative(Neutral) | Total |
|------------|----------------|-------------------|-------|
| Training   | 28930          | 27984             | 56914 |
| Validation | 6092           | 6104              | 12196 |
| Testing    | 6132           | 6064              | 12196 |
| Total      | 41154          | 40152             | 81306 |

We observe from Fig. 3 that the accuracy increases in a steady manner during training while the validation accuracy fluctuates around 96%. The best result at 96.43% is achieved at epoch 50.

We observe that the CNN-Mixture model gains small but constant accuracy improvement in both validation and testing sets, where it achieves 96.88% in validation set and 96.90% testing set, respectively as shown in Table 2. From the experimental results, we find that CNN-Mixture can effectively distinguish an adult image from a neutral image. Moreover, CNN-Mixture outperforms the single CNN model and even the average sum of multiple CNNs. Compared with the average sum, the CNN-Mixture uses the OLS to find the proper contribution for each model which help to distinguish better the role of each model in extracting useful features for recognizing the image content.

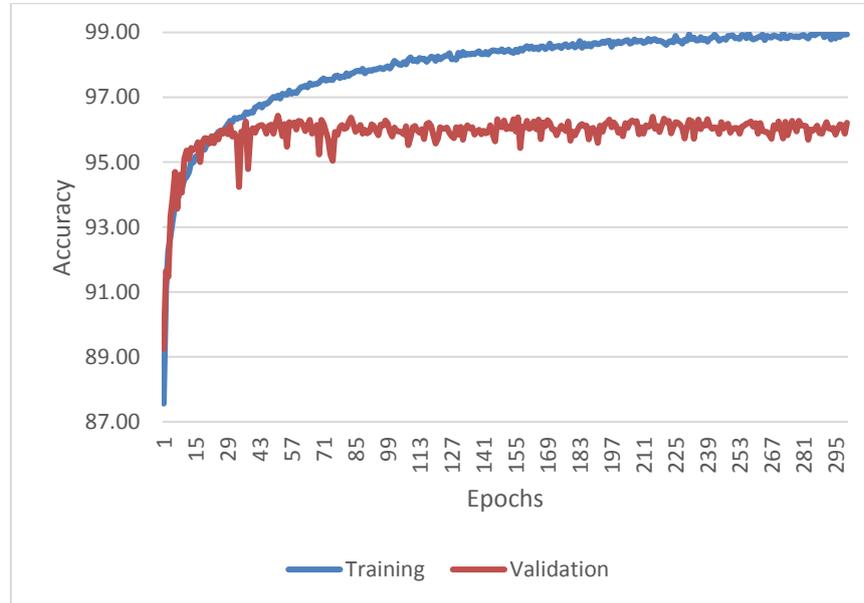

**Fig. 3.** The performance of the model during training

**Table 2.** Comparison of the models based on accuracy

|            | CNN-Mixture | Average Sum | Single CNN |
|------------|-------------|-------------|------------|
| Validation | **96.88%**  | 96.44%      | 96.43%     |
| Testing    | **96.90%**  | 96.50%      | 96.34%     |

## 6    Conclusion

In this paper, we propose an adult image recognition system using a mixture of CNN. The proposed model is end-to-end learnable as compared to traditional solutions such as skin color detection. The proposed method is formed as a linear regression problem where the weights are calculated using ordinary least square. The proposed model is tested on a manually collected large dataset consisting of over hundred thousand of training images. Experiment results show that CNN-Mixture is effective against recognition of adult images, yielding an accuracy of over 96% on the testing set.